\documentclass[10pt,twocolumn,letterpaper]{article}

\usepackage[pagenumbers]{cvpr}      
\usepackage{float}
\definecolor{cvprblue}{rgb}{0.21,0.49,0.74}
\usepackage[pagebackref,breaklinks,colorlinks,allcolors=cvprblue]{hyperref}

\def\paperID{*****} 
\def\confName{CVPR}
\def\confYear{2026}

\title{Expanding the Content-Style Frontier: a Balanced Subspace Blending Approach for Content-Style LoRA Fusion}

\author{Linhao Huang\\
Beijing University of Technology \\
{\tt\small huanglinhao@emails.bjut.edu.cn}
}

\begin{document}
\twocolumn[{%
\maketitle
\begin{figure}[H]
\hsize=\textwidth 
  \centering
  \begin{subfigure}{1.2\linewidth}
    \includegraphics[width=0.95\linewidth]{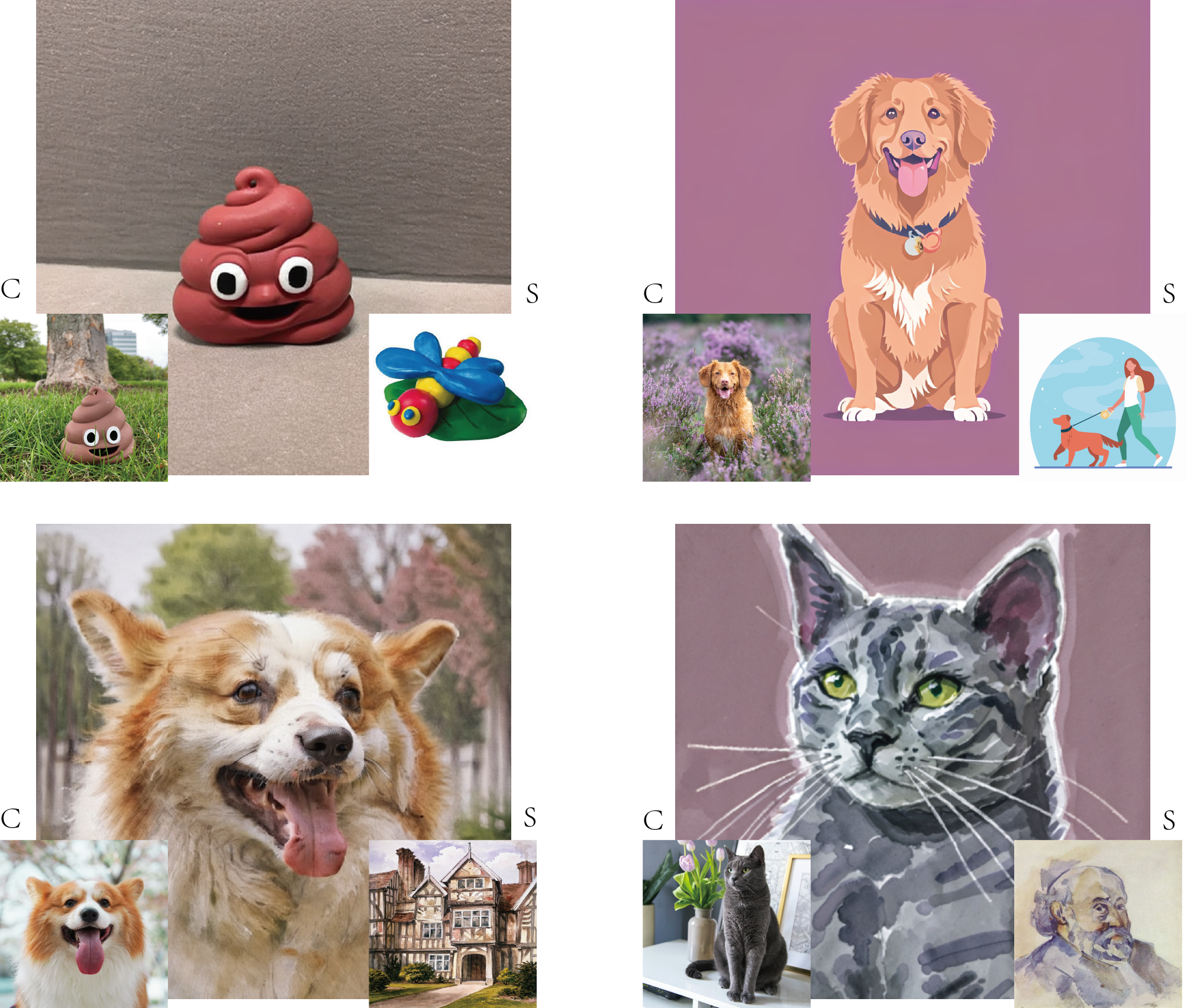}
    \caption{}
    \label{fig:short-a}
  \end{subfigure}
  \begin{subfigure}{0.7\linewidth}
    \includegraphics[width=0.95\linewidth]{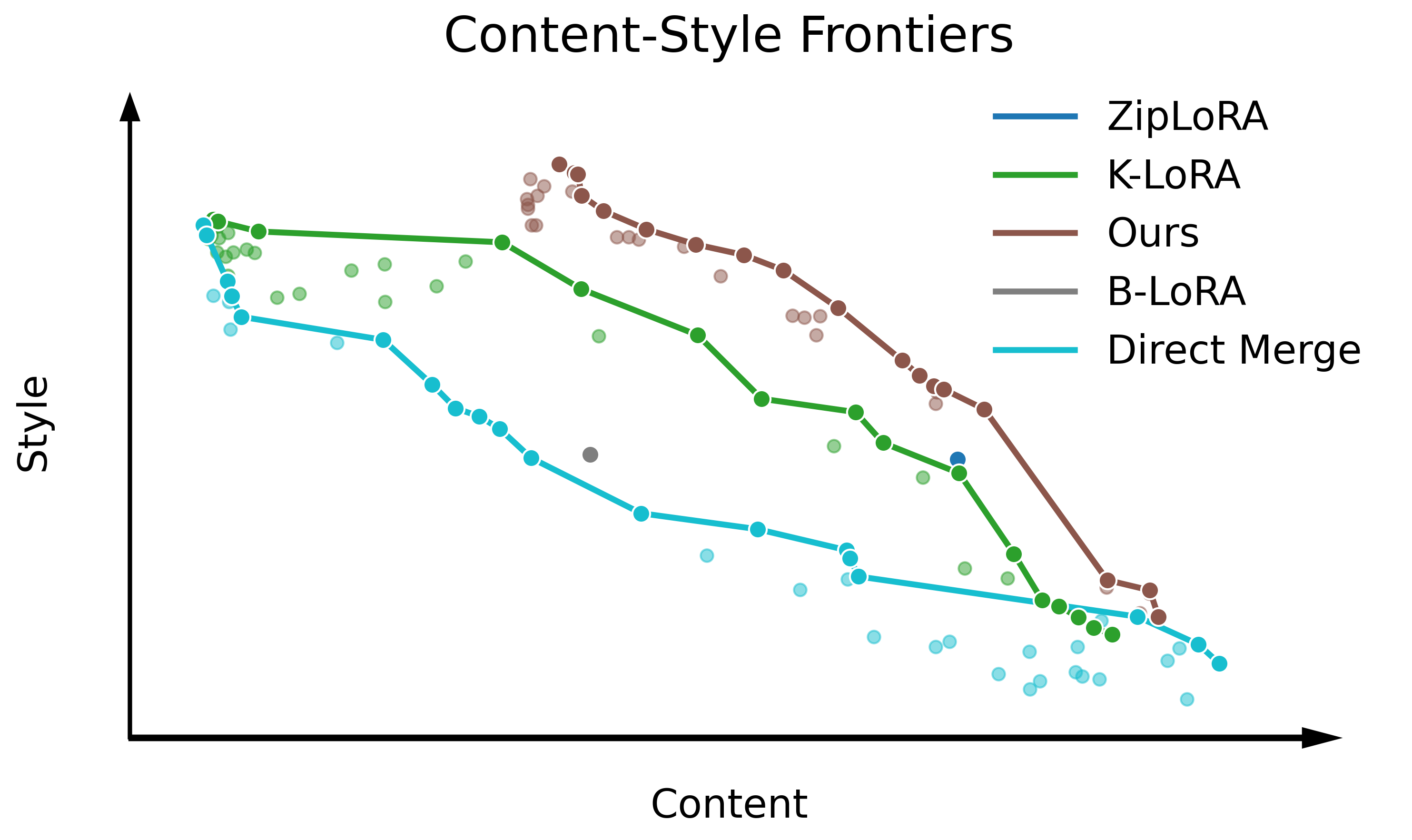}

   \caption{}
   \label{fig:frontier}
  \end{subfigure}
  \caption{In our experiments with existing models, we observed that these models either lack the ability to adjust style intensity or suffer from a significant loss of content features as the style intensity increases. Consequently, this results in a suboptimal content–style frontier. In contrast, our objective is to expand this frontier.}
  \label{fig:short}
\end{figure}

}]
\begin{abstract}
Recent advancements in text-to-image diffusion models have significantly improved the personalization and stylization of generated images. However, previous studies have only assessed content similarity under a single style intensity. In our experiments, we observe that increasing style intensity leads to a significant loss of content features, resulting in a suboptimal content-style frontier. To address this, we propose a novel approach to expand the content-style frontier by leveraging Content-Style Subspace Blending and a Content-Style Balance loss. Our method improves content similarity across varying style intensities, significantly broadening the content-style frontier. Extensive experiments demonstrate that our approach outperforms existing techniques in both qualitative and quantitative evaluations, achieving superior content-style trade-off with significantly lower Inverted Generational Distance (IGD) and Generational Distance (GD) scores compared to current methods. 
\end{abstract}
\section{Introduction}
Personalization and stylization have long been central topics in computer vision, aiming to adapt image generation models to user-specific concepts or stylistic preferences. Personalization typically focuses on capturing distinct content, such as specific objects or subjects, while stylization emphasizes transferring visual attributes like color, texture, and artistic patterns. Despite their importance, achieving controllable content-style manipulation remains challenging due to the strong entanglement between these two factors.

Recent advances in text-to-image diffusion models have significantly expanded the scope of personalization and stylization. Early approaches such as Textual Inversion\cite{Textual_Inversion} and DreamBooth\cite{DreamBooth} enable subject-specific adaptation, but either suffer from limited expressiveness or require costly model fine-tuning. More efficient parameterization techniques, such as LoRA\cite{lora} and its extensions\cite{MoSLoRA}, have further popularized diffusion-based customization by allowing lightweight fine-tuning. This has motivated a growing body of work on combining content and style LoRAs, aiming to flexibly render arbitrary subjects in arbitrary styles. Methods such as ZipLoRA\cite{ZipLoRA} and Multi-LoRA Composition\cite{LoRA_Composition} merge or sequentially activate LoRAs to achieve subject-style fusion, while more advanced techniques like K-LoRA\cite{K-LoRA} adjust the timing of LoRA application based on a Top-K selection mechanism during the denoising process to balance style fidelity and content preservation. Beyond fusion, approaches like B-LoRA\cite{B-LoRA} and DuoLoRA\cite{DuoLoRA} explicitly pursue disentanglement of content and style, respectively by training different blocks and enforcing cycle-consistency across content and style.

In practice, users exhibit diverse preferences for style intensity, which can also vary depending on the specific combination of content and style. However, previous studies have evaluated content similarity only under a single level of style intensity. In our experiments with existing models, we observed that these models either lack the ability to adjust style intensity or suffer from a significant loss of content features as the style intensity increases. Consequently, this results in a suboptimal content–style frontier, as illustrated in \cref{fig:frontier}. In contrast, our objective is to expand this frontier. 

To address this issue, during the training phase, we introduce Content-Style Subspace Blending by learning mixing matrices to flexibly merge the content and style LoRA subspaces. To optimize this process, we design the Content-Style Balance loss. Comprising a loss-balancing regularization component, this loss ensures that intermediate states generated by the blending weight retain at least one of the key features (content or style), thus preventing the simultaneous feature degradation observed in Direct Merge technique. During the inference phase, we employ a Non-linear Content-Style Blending strategy, leveraging the property of diffusion models to generate content structure early and style details later. 

The main contributions of our work are summarized as follows:

\begin{itemize}
    \item We propose a novel Content-Style Subspace Blending method that utilizes learnable parameters to fuse content and style LoRA subspaces, offering superior flexibility and performance compared to simple LoRA arithmetic merging.
    \item We introduce the Content-Style Balance loss, which explicitly regularizes the intermediate states between content and style, effectively solving the problem of simultaneous feature degradation that occurs in Direct Merge technique.
    \item We devise a time-dependent, Non-linear Content-Style Blending strategy that leverages the inherent properties of diffusion models, generating content structure early and style details later. This approach leads to significant improvements in both content preservation and style expression.
    \item  Extensive experiments demonstrate that our method significantly expands the content-style frontier. Quantitative experiments show that our approach surpasses baselines in both IGD and GD metrics.
\end{itemize}

\section{Related work}

\textbf{Diffusion Models for Personalization.} Recent advancements in text-to-image diffusion models have sparked significant interest in techniques for model personalization and adaptation. Textual Inversion \cite{Textual_Inversion} proposes a straightforward approach that learns a new token embedding to represent a specific concept, enabling the model to generate images consistent with user-provided examples. While this method is efficient, it is limited by the expressive capacity of the frozen diffusion model. To address these limitations, DreamBooth \cite{DreamBooth} fine-tunes the entire diffusion model on subject-specific images, achieving high-fidelity personalization at the cost of increased training requirements and a potential risk of overfitting.

Building on these ideas, Custom Diffusion \cite{Custom_Diffusion} introduces a concept customization framework that selectively updates only the key and value projection matrices corresponding to the text features. This selective fine-tuning allows for the generation of images with multiple customized concepts through an optimization-based approach. Similarly, StyleDrop \cite{StyleDrop} focuses on style adaptation, enabling the transfer of arbitrary artistic styles through efficient fine-tuning techniques. In parallel, Low-Rank Adaptation (LoRA) \cite{lora} presents a parameter-efficient fine-tuning strategy, which was later extended by MoSLoRA \cite{MoSLoRA}, introducing a trainable mixer that combines the subspaces of LoRA.

\textbf{LoRA Combination for Image Stylization.} A growing body of research has focused on combining LoRAs to enable flexible subject–style fusion for image stylization. ZipLoRA \cite{ZipLoRA} introduces an effective merging strategy that facilitates the rendering of arbitrary subjects in arbitrary styles by aligning and integrating pairs of subject and style LoRAs. Multi-LoRA Composition \cite{LoRA_Composition} extends this line of work by presenting two training-free methods: sequentially activating individual LoRAs and conditioning the classifier-free guidance\cite{CFG} (CFG) on the LoRAs. More recently, K-LoRA \cite{K-LoRA} proposes a training-free approach, demonstrating that applying style LoRAs in later timesteps preserves style fidelity, while applying content LoRAs in earlier timesteps yields significantly improved results.

Beyond fusion techniques, some methods explicitly aim to disentangle content and style, enhancing control over the generated images. B-LoRA \cite{B-LoRA} seeks implicit content–style separation by training two layers within the base text-to-image model: one layer capturing content and the other capturing style. DuoLoRA \cite{DuoLoRA} introduces ZipRank, a mechanism that merges content and style along the rank dimension, offering adaptive rank flexibility while significantly reducing the number of learnable parameters.

\section{Method}
\begin{figure*}
  \centering
   \includegraphics[width=0.95\linewidth]{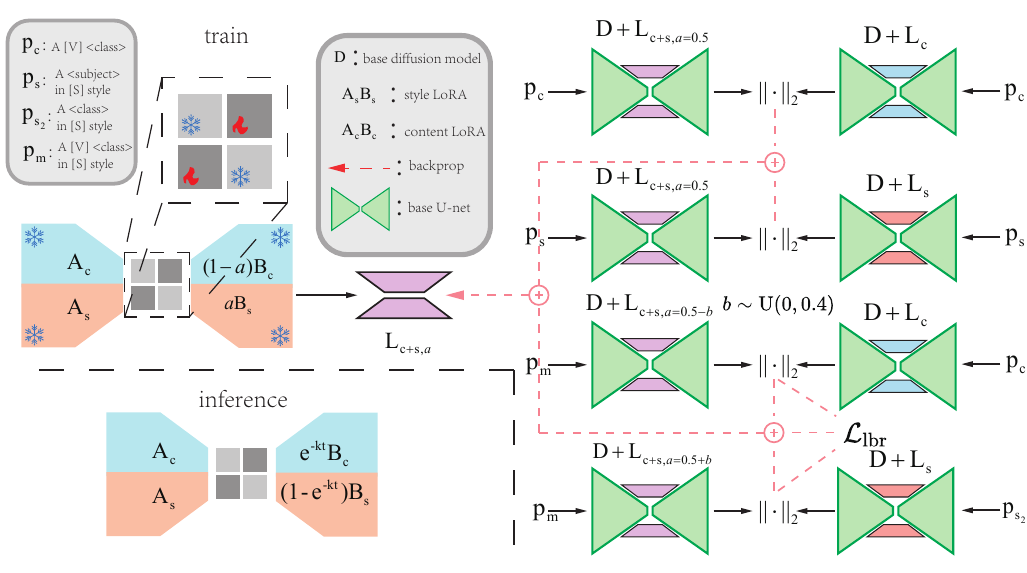}

   \caption{\textbf{Overview of our method.} Our approach consists of two stages: training and inference. In the train phase, we introduce Content-Style Subspace Blending, which utilizes learnable mixing matrices to fuse the content ($\mathbf{A}_\text{c}, \mathbf{B}_\text{c}$) and style ($\mathbf{A}_\text{s}, \mathbf{B}_\text{s}$) LoRA subspaces. This process is guided by our proposed Content-Style Balance loss. In the inference phase, we apply a Non-linear Content-Style Blending strategy, which uses dynamic, time-dependent weights during the denoising process. This approach leverages the property of diffusion models to generate content structure early and style details later, thereby achieving a superior content-style trade-off.}
   \label{fig:framework}
\end{figure*}
\subsection{Preliminaries}
Low-Rank Adaptation (LoRA) \cite{lora} is a parameter-efficient fine-tuning technique initially proposed for adapting large-scale language models and has since been widely adopted in vision and diffusion models. The core idea behind LoRA is that the weight updates introduced during fine-tuning, denoted as $\Delta \mathbf{W} \in \mathbb{R}^{m \times n}$, often exhibit a low-rank structure. Rather than directly optimizing $\Delta \mathbf{W}$, LoRA parameterizes it as the product of two low-rank matrices:
\begin{equation}
\Delta \mathbf{W} = \mathbf{A}\mathbf{B}, \quad \mathbf{A} \in \mathbb{R}^{m \times r}, \quad \mathbf{B} \in \mathbb{R}^{r \times n}, \quad r \ll \min(m,n).
\end{equation}
Here, $r$ represents the intrinsic rank, which controls the capacity for adaptation. During training, the original model weights $\mathbf{W}_0$ remain frozen, while only the matrices $\mathbf{A}$ and $\mathbf{B}$ are updated. At inference, the effective weights are computed as:
\begin{equation}
\mathbf{W} = \mathbf{W}_0 + \Delta \mathbf{W} = \mathbf{W}_0 + \mathbf{A}\mathbf{B}.
\end{equation}
This factorization dramatically reduces the number of trainable parameters while preserving the expressiveness of the adapted model, making LoRA a practical and efficient approach for large-scale model adaptation.

\subsection{Non-linear content-style blending}
To expand the content–style frontier, we introduce a Non-linear Content–Style Blending strategy. Specifically, given two independently trained LoRA weights, $\Delta \mathbf{W}_\text{c}=\mathbf{A}_\text{c}\mathbf{B}_\text{c}$ and $\Delta \mathbf{W}_\text{s}=\mathbf{A}_\text{s}\mathbf{B}_\text{s}$, representing the content and style LoRA respectively, we mix them as follows:
\begin{equation}
    \Delta \mathbf{W} =e^{-kt}\mathbf{A}_\text{c}\mathbf{B}_\text{c}+(1-e^{-kt})\mathbf{A}_\text{s}\mathbf{B}_\text{s},
\end{equation}
where $t$ denotes the current step in the backward denoising process and ranges $[0, 1]$. The hyperparameter $k$ controls the rate of content decay. At the start of the denoising process ($t = 0$), we have $\Delta \mathbf{W}=\mathbf{A}_\text{c}\mathbf{B}_\text{c}$. As denoising progresses, $t$ gradually increases, allowing the style to emerge. When the process ends ($t = 1$), we obtain $\Delta \mathbf{W} =e^{-k}\mathbf{A}_\text{c}\mathbf{B}_\text{c}+(1-e^{-k})\mathbf{A}_\text{s}\mathbf{B}_\text{s}$. For sufficiently large $k$, $e^{-k}\approx0$, leading to $\Delta \mathbf{W}\approx \mathbf{A}_\text{s}\mathbf{B}_\text{s}$. 

This approach fully leverages the insights from \cite{K-LoRA,Freetuner}, which suggest that diffusion models generate images in a coarse-to-fine manner: they first form the overall content and then refine the fine-grained details. Accordingly, applying the content LoRA in earlier timesteps and the style LoRA in later timesteps helps preserve stylistic information while avoiding interference with the original object structure. Unlike K-LoRA, our method supports continuous intermediate states between content and style, enabling more precise blending control.

\begin{figure}[t]
  \centering
  
   \includegraphics[width=\linewidth]{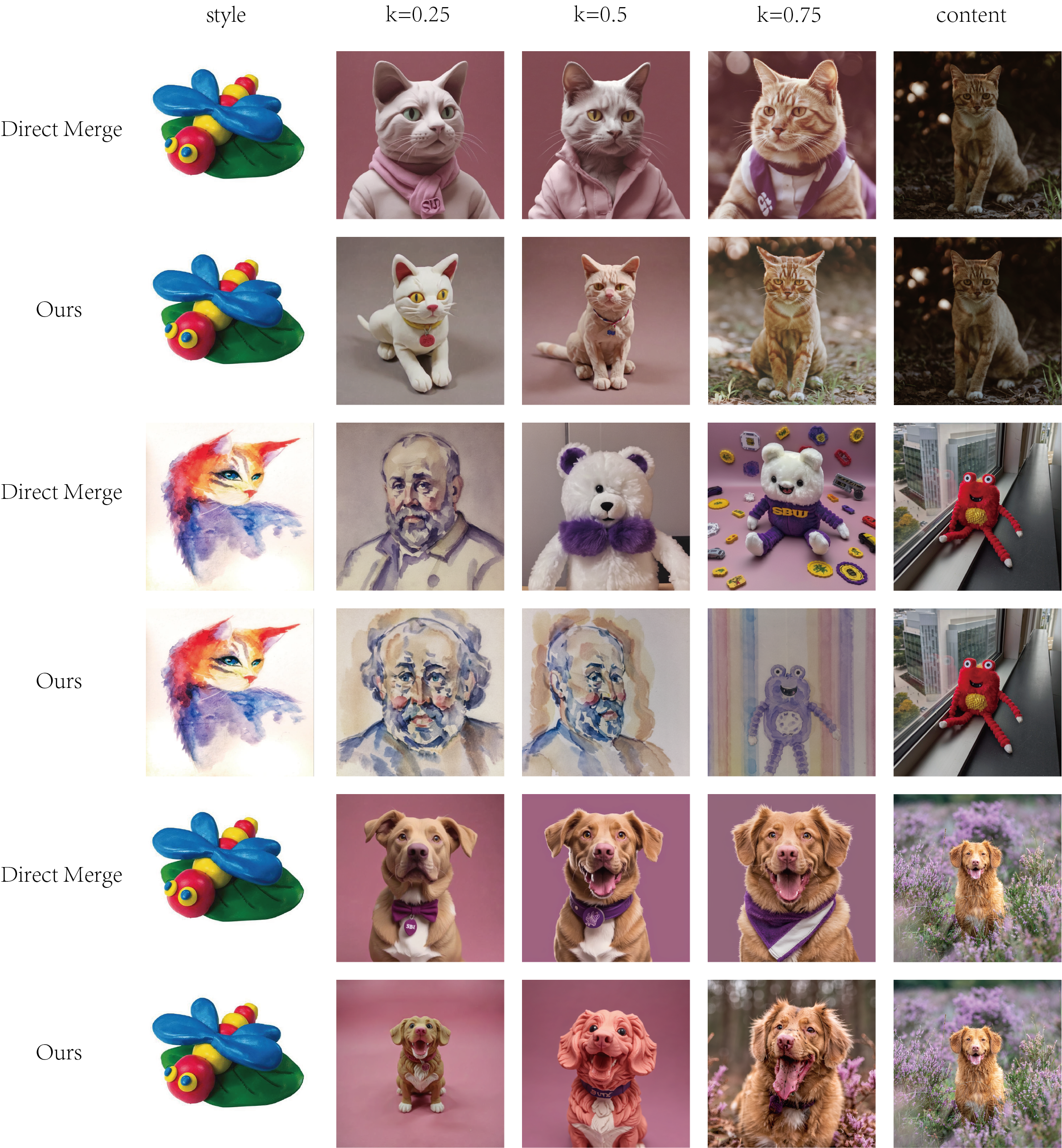}

   \caption{The intermediate states produced by the Direct Merge method often lose both content and style characteristics. In contrast, our method ensures that the intermediate states retain at least one of either the content or the style features. To ensure a fair comparison, we do not employ Non-linear Content-Style Blending, but instead utilize a weighted summation method akin to Direct Merge.}
   \label{fig:ob}
\end{figure}

However, we observed that the performance gain obtained from using this non-linear approach alone was limited. To further investigate, we introduced the weight $k$ in the Direct Merge method to explore intermediate states between content and style:
\begin{equation}
    \triangle \mathbf{W}=k\mathbf{A}_\text{c} \mathbf{B}_\text{c}+(1-k)\mathbf{A}_\text{s}\mathbf{B}_\text{s}\text{,}
    \label{Direct Merge}
\end{equation}
and the results are illustrated in \cref{fig:ob}. We found that the intermediate states produced by the Direct Merge method often lose both content and style characteristics simultaneously, leading to a suboptimal content–style trade-off frontier. To address this issue, we propose a Content–Style Balance loss, which ensures that intermediate states retain at least one of the two key feature types—content or style—as shown in \cref{fig:ob}. Furthermore, we introduce Content-Style Subspace Blending to train the model.

\label{sec:Non-linear Content-Style Blending}

\subsection{Subspace perspective}
Given two independently trained LoRA weights, $\Delta \mathbf{W}_\text{c}=\mathbf{A}_\text{c}\mathbf{B}_\text{c}$ and $\Delta \mathbf{W}_\text{s}=\mathbf{A}_\text{s}\mathbf{B}_\text{s}$, which represent the content and style LoRAs respectively, we can view the combination of content LoRA and style LoRA as a new LoRA, expressed as:
\begin{equation}
    \Delta \mathbf{W} =\mathbf{A}_\text{c}\mathbf{B}_\text{c}+\mathbf{A}_\text{s}\mathbf{B}_\text{s}=
\begin{bmatrix}
\mathbf{A}_\text{c} & \mathbf{A}_\text{s}
\end{bmatrix}
\begin{bmatrix}
\mathbf{B}_\text{c} \\
\mathbf{B}_\text{s}
\end{bmatrix}
=\mathbf{A}_\text{m}\mathbf{B}_\text{m},
\label{subsapce}
\end{equation}
where $\mathbf{A}_\text{m}=\begin{bmatrix}
\mathbf{A}_\text{c} & \mathbf{A}_\text{s}
\end{bmatrix}\in \mathbb{R}^{m \times 2r},\mathbf{B}_\text{m}=\begin{bmatrix}
\mathbf{B}_\text{c} \\
\mathbf{B}_\text{s}
\end{bmatrix}\in \mathbb{R}^{2r \times n}$. From this perspective, previous works\cite{ZipLoRA,DuoLoRA} can be interpreted as different methods of summing the subspaces\cite{MoSLoRA} of $\mathbf{A}_\text{m}\mathbf{B}_\text{m}$. In this context, subspaces are defined as parallel components with lower rank values. For example, in \cref{subsapce}, $\mathbf{A}_\text{c}\mathbf{B}_\text{c}$ and $\mathbf{A}_\text{s}\mathbf{B}_\text{s}$ are two subspaces of $\mathbf{A}_\text{m}\mathbf{B}_\text{m}$. In ZipLoRA\cite{ZipLoRA}, the combination of the two LoRAs is formulated as:
\begin{equation}
\begin{split}
    \Delta \mathbf{W} &= m_\text{c} \otimes \Delta \mathbf{W}_\text{c} + m_\text{s} \otimes \Delta \mathbf{W}_\text{s}\\
    &=
\begin{bmatrix}
\mathbf{A}_\text{c} & \mathbf{A}_\text{s}
\end{bmatrix}
\begin{bmatrix}
\mathbf{B}_\text{c}\text{diag}(m_\text{c}) \\
\mathbf{B}_\text{s}\text{diag}(m_\text{s})
\end{bmatrix},
\end{split}
\end{equation}
 where $m_\text{c},m_\text{s}\in \mathbb{R}^{1 \times n}$, and $\otimes$ denotes element-wise multiplication between $\Delta \mathbf{W}$ and the broadcasted merger coefficient vector $m$. Specifically, the $j^{\text{th}}$ column of $\Delta \mathbf{W}$ is multiplied by the $j^{\text{th}}$ element of $m$. ZipLoRA defines a loss function that includes the term $\sum_{i} \left| m_\text{c}^{(i)} \cdot m_\text{s}^{(i)} \right|$, which aims to minimize signal interference when merging the two subspaces. This ensures that when the columns of the subspaces are combined, interference between the signals is minimized as much as possible. In DuoLoRA\cite{DuoLoRA}, the combination is expressed as:
\begin{equation}
\begin{split}
    \Delta \mathbf{W} &= \mathbf{A}_\text{c}\text{diag}(m_\text{c}^\prime)\mathbf{B}_\text{c}+\mathbf{A}_\text{s}\text{diag}(m_\text{s}^\prime)\mathbf{B}_\text{s}\\
    &=
\begin{bmatrix}
\mathbf{A}_\text{c} & \mathbf{A}_\text{s}
\end{bmatrix}
\begin{bmatrix}
\text{diag}(m_\text{c}^\prime) & \mathbf{O}\\
\mathbf{O} & \text{diag}(m_\text{s}^\prime)
\end{bmatrix}
\begin{bmatrix}
\mathbf{B}_\text{c} \\
\mathbf{B}_\text{s}
\end{bmatrix},
\end{split}
\end{equation}
where $m_c^\prime,m_s^\prime \in \mathbb{R}^{1 \times r}$. $\mathbf{O}$ denotes the zero matrix. Finally, the naive weighted sum method can be interpreted as a simple weighted sum of the two subspaces:
\begin{equation}
\begin{split}
    \Delta \mathbf{W} &= 
(1-a)\Delta \mathbf{W}_\text{c} + a\Delta \mathbf{W}_\text{s}\\
&=
\begin{bmatrix}
\mathbf{A}_\text{c} & \mathbf{A}_\text{s}
\end{bmatrix}
\begin{bmatrix}
(1-a)\mathbf{B}_\text{c} \\
a\mathbf{B}_\text{s}
\end{bmatrix}.
\end{split}
\end{equation}

\subsection{Content-style subspace blending}
While previous work has explored various methods for combining these two subspaces, we observe that they have overlooked the potential benefits of blending them. MoSLoRA\cite{MoSLoRA} has already demonstrated that blending subspaces within LoRA improves performance, and in this work, we extend that idea by introducing Content-Style Subspace Blending. Specifically, our method defines the update to the weight matrix as:
\begin{equation}
\begin{split}
    \Delta \mathbf{W}
&=
\begin{bmatrix}
\mathbf{A}_\text{c} & \mathbf{A}_\text{s}
\end{bmatrix}
\begin{bmatrix}
\mathbf{I} & \mathbf{W}_{12}\\
\mathbf{W}_{21} & \mathbf{I}
\end{bmatrix}
\begin{bmatrix}
\mathbf{B}_\text{c} \\
\mathbf{B}_\text{s}
\end{bmatrix}\\
&=\mathbf{A}_\text{c}\mathbf{B}_\text{c}+\mathbf{A}_\text{s}\mathbf{B}_\text{s}+\mathbf{A}_\text{s}\mathbf{W}_{21}\mathbf{B}_\text{c}+\mathbf{A}_\text{c}\mathbf{W}_{12}\mathbf{B}_\text{s}\text{.}
\end{split}
\label{Content-Style Subspace Blending}
\end{equation}
Here, the learnable parameters $\mathbf{W}_{12}$ and $\mathbf{W}_{21}$ are initialized as zero matrices of size $r \times r$, and $\mathbf{I}$ denotes the identity matrix. It is important to note that we do not blend the subspaces of $\mathbf{A}_\text{c}\mathbf{B}_\text{c}$ and $\mathbf{A}_\text{s}\mathbf{B}_\text{s}$ themselves. This approach preserves the original performance of content and style LoRA components, while offering enhanced flexibility. Our ablation experiments show that this method outperforms other approaches that blend the subspaces of $\mathbf{A}_\text{m}\mathbf{B}_\text{m}$.

In addition, building on the observations from \cref{sec:Non-linear Content-Style Blending}, we propose the Content-Style Balance loss to guide training.
\subsection{Content-style balance loss}

To capture intermediate states between content and style, we introduce a blending parameter $a$, as shown in \cref{Content-Style Subspace Blending with a}.
\begin{equation}
\begin{split}
    \Delta \mathbf{W}
&=
\begin{bmatrix}
\mathbf{A}_\text{c} & \mathbf{A}_\text{s}
\end{bmatrix}
\begin{bmatrix}
\mathbf{I} & \mathbf{W}_{12}\\
\mathbf{W}_{21} & \mathbf{I}
\end{bmatrix}
\begin{bmatrix}
(1-a)\mathbf{B}_\text{c} \\
a\mathbf{B}_\text{s}
\end{bmatrix}\\
&=(1-a)\mathbf{A}_\text{c}\mathbf{B}_\text{c}+a\mathbf{A}_\text{s}\mathbf{B}_\text{s}\\
&\quad+(1-a)\mathbf{A}_\text{s}\mathbf{W}_{21}\mathbf{B}_\text{c}+a\mathbf{A}_\text{c}\mathbf{W}_{12}\mathbf{B}_\text{s}\text{.}
\end{split}
\label{Content-Style Subspace Blending with a}
\end{equation}
Suppose the text conditioning prompts for the content and style references are defined as $p_\text{c}=$"A [V] $\textless$class$\textgreater$" and $p_\text{s}=$"A $\textless$subject$\textgreater$ in [S] style", respectively. We then construct two additional prompts: $p_{\text{s}_2}=$"A $\textless$class$\textgreater$ in [S] style" and $p_\text{m}=$"A [V] $\textless$class$\textgreater$ in [S] style". Based on these definitions, we propose a merger loss $\mathcal{L}_\text{merger}$ to ensure that, when $a$ takes values within $[0, 1]$, the images generated by the model exhibit at least one of the desired content or style characteristics:
\begin{equation}
\begin{split}
    \mathcal{L}_\text{merger}&=\left\| \left( D \oplus L_{\text{m},a=0.5-b} \right)(x_\text{c}, p_\text{m}) - \left( D \oplus L_\text{c} \right)(x_\text{c}, p_\text{c}) \right\|_2\\
    &+\left\| \left( D \oplus L_{\text{m},a=0.5+b} \right)(x_\text{s}, p_\text{m}) - \left( D \oplus L_\text{s} \right)(x_\text{s}, p_{\text{s}_2}) \right\|_2
\end{split}\textbf{,}
\end{equation}
where $(x_\text{c}, x_\text{s})$ denote the noisy latents of the content and style references, respectively. $L_\text{c}$ and $L_\text{s}$ represent the content and style LoRAs, $D$ is base diffusion model, and $L_\text{m}$ is our proposed blended LoRA. Before each loss computation, we sample $b \sim \rm{U}(0, 0.4)$, where $\rm{U}(\cdot)$ denotes a uniform distribution. 

Intuitively, minimizing $\mathcal{L}_\text{merger}$ encourages the model to generate images that emphasize content features when $a \in [0, 0.5]$, and style features when $a\in [0.5, 1]$. However, we observed in practice that the model may sometimes minimize only one component of this loss. To address this issue, we introduce a loss-balancing regularization term that prevents imbalance between the two loss components:
\begin{equation}
\begin{split}
\mathcal{L}_\text{lbr}&=\text{ReLU}(\mathcal{L}_\text{c}-\mathcal{L}_\text{s}.\text{detach}())^2\\
&+\text{ReLU}(\mathcal{L}_\text{s}-\mathcal{L}_\text{c}.\text{detach}())^2\text{,}
\end{split}
\end{equation}
where $\mathcal{L}_\text{c}$ and $\mathcal{L}_\text{s}$ correspond to the two components of $\mathcal{L}_\text{merger}$. This regularization penalizes the inconsistency between the two losses, preventing the model from optimizing a single direction excessively. 

To further preserve the original behavior of both the content and style LoRAs within the merged model, we propose a prior-preserving loss:
\begin{equation}
\begin{split}
\mathcal{L}_\text{prior}&=\left\| \left( D \oplus L_{\text{m},a=0.5} \right)(x_\text{c}, p_\text{c}) - \left( D \oplus L_\text{c} \right)(x_\text{c}, p_\text{c}) \right\|_2\\
&+\left\| \left( D \oplus L_{\text{m},a=0.5} \right)(x_\text{s}, p_\text{s}) - \left( D \oplus L_\text{s} \right)(x_\text{s}, p_\text{s}) \right\|_2\text{,}
\end{split}
\end{equation}

Finally, our Content-Style Balance loss is defined as:
\begin{equation}
\mathcal{L}_\text{csb}=\mathcal{L}_\text{prior}+\lambda_\text{merger}\mathcal{L}_\text{merger}+\lambda_\text{lbr}\mathcal{L}_\text{lbr}\text{,}
\end{equation}
where $\lambda_\text{merger}$ and $\lambda_\text{lbr}$ are hyperparameters that control the relative contribution of each component.

During the inference stage, we apply our proposed Non-linear Content-Style Blending. Formally, 
\begin{equation}
\begin{split}
    \Delta \mathbf{W}
&=
\begin{bmatrix}
\mathbf{A}_\text{c} & \mathbf{A}_\text{s}
\end{bmatrix}
\begin{bmatrix}
\mathbf{I} & \mathbf{W}_{12}\\
\mathbf{W}_{21} & \mathbf{I}
\end{bmatrix}
\begin{bmatrix}
e^{-kt}\mathbf{B}_\text{c} \\
(1-e^{-kt})\mathbf{B}_\text{s}
\end{bmatrix}
\end{split}
\label{Non-linear Content-Style Blending}
\end{equation}

\section{Experiments}
\subsection{Experiment setup}

\textbf{Datasets.} For the content LoRA, we use a diverse set of content images from the DreamBooth\cite{DreamBooth} dataset. Previous studies\cite{CustomizingText-to-ImageModelswithaSingleImagePair,StyleDrop} have noted that training style-tuned model using only a single style image often causes the model to replicate content rather than the intended style. To mitigate this issue, we select our style data from the StyleBench\cite{StyleShot}, where each style category contains 3–5 representative images. 

\noindent\textbf{Experimental details.} All experiments are conducted using SDXL v1.0 as the base diffusion model. Both content and style LoRAs are trained using DreamBooth fine-tuning with a LoRA rank of 64. We optimize the LoRA parameters for 1,000 steps using the AdamW optimizer, with a batch size of 1 and a learning rate of $5\times10^{-5}$. The text encoder of SDXL remains frozen during the LoRA finetuning.

During the merging, we train the merged LoRA for 300 steps using the proposed Content-Style Balance loss. The optimizer is AdamW with a learning rate of $1\times10^{-3}$. The hyperparameters are set to $\lambda_\text{merger}=0.8$ and $\lambda_\text{lbr}=1$.

\subsection{Quantitative and qualitative results}

We randomly select 10 combinations from 7 objects and 5 styles to perform quantitative evaluations. Our method is compared with several widely adopted and state-of-the-art approaches, including direct arithmetic merging, ZipLoRA\cite{ZipLoRA}, B-LoRA\cite{B-LoRA} and K-LoRA\cite{K-LoRA}. To evaluate performance, we employ CLIP\cite{CLIP} similarity to measure style alignment and compute the subject similarity through DINO\cite{DINO} score. By varying each model’s hyperparameters, we obtain multiple pairs of (subject similarity, style similarity) values. 

Specifically, for K-LoRA, the parameter $\beta$ in its Equation (7) is varied over the range $[0, 2]$ with a step size of 0.05. For Direct Merge (see \cref{Direct Merge}), the coefficient $k$ is varied over $[0, 1]$ with a step size of 0.025. For our method, $k$ is varied over $[0, 10]$ with a step size of 0.25. For B-LoRA and ZipLoRA, only a single data point can be obtained since these methods do not allow continuous control of style strength during inference.

The Pareto frontier represents the set of optimal trade-off solutions where one objective cannot be improved without degrading the other. To evaluate model performance, we plot the content–style frontier and compute two standard multi-objective metrics: the Inverted Generational Distance (IGD) and the Generational Distance (GD). IGD measures the average distance from the true Pareto front to the model’s front, reflecting diversity and coverage, while GD measures the average distance in the opposite direction, reflecting approximation quality.

Formally, let $P$ denote the Pareto front obtained by a model, and $P^*$ the combined front of all methods. The two metrics are defined as: 
\begin{equation}
    \text{IGD} = \frac{1}{|P^*|} \sum_{\substack{x \in P^* }} \min_{y \in P} d(x, y)\text{,}
\end{equation}
\begin{equation}
    \text{GD} = \frac{1}{|P|} \sum_{\substack{y \in P}} \min_{x \in P^*} d(y,x)\text{,}
\end{equation}
where $d(x, y)$ denotes the Euclidean distance between two points. A model's IGD and GD values may vary across different experiments, as the set $P^*$ changes between experiments.

\begin{figure}[htbp]
  \centering
  \begin{subtable}{0.48\textwidth}
    \centering
    \caption{Comparison of alignment results across different methods.}
    \begin{tabular}{@{}ccc@{}}
    \toprule
    Method& IGD$\downarrow$& GD$\downarrow$\\
    \midrule
     Direct Merge& 0.0626& 0.111\\
     B-LoRA& 0.2114&0.068\\
     ZipLoRA& 0.1468&0.0215 \\
     K-LoRA& 0.0323&0.0627 \\
     Ours& \textbf{0.0055}&\textbf{0}\\
    \bottomrule
    \end{tabular}
    \label{tab:diff_methods}
  \end{subtable}\hfill
  \begin{subfigure}{0.48\textwidth}
    \centering
    \includegraphics[width=0.95\linewidth]{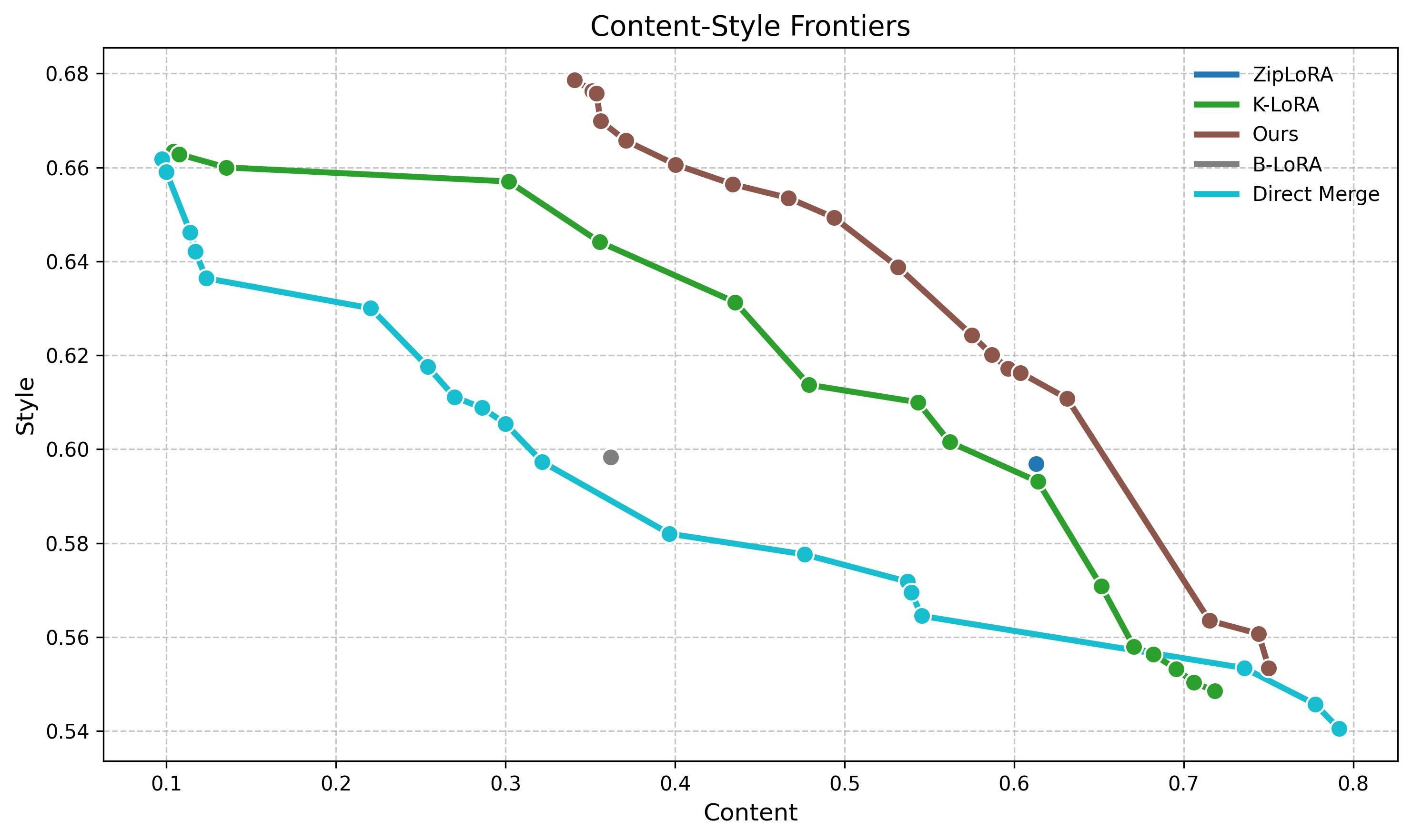}
    \caption{Content-style frontier of different methods. For a clearer comparison, we only show content-style frontier and omitted points that are not at the frontier of each method.}
    \label{fig:diff_methods}
  \end{subfigure}
\caption{\textbf{Quantitative comparisons.} Comparison of alignment results across different methods.}
\end{figure}

The quantitative results are summarized in \cref{tab:diff_methods}. Our method achieves the lowest IGD and GD scores among all approaches, indicating both superior coverage and closer approximation to the optimal content-style trade-off frontier. These results demonstrate that our approach significantly extends the achievable frontier, outperforming previous methods in balancing content preservation and style expression.

\begin{figure*}
  \centering
   \includegraphics[width=0.95\linewidth]{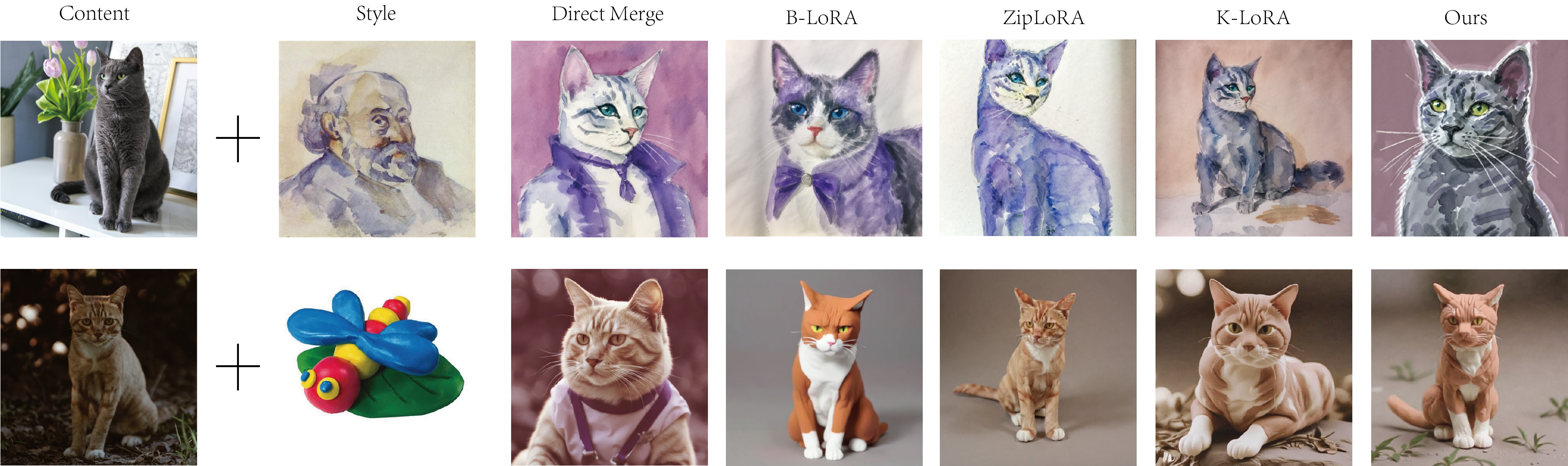}

   \caption{\textbf{Qualitative comparisons.} We present images generated by our method and the compared methods.}
   \label{fig:qualitative}
\end{figure*}

Qualitative comparison results are presented in \cref{fig:qualitative}. As shown, our method clearly outperforms the baselines, as evidenced by the visual results.

\subsection{Ablation studies}
\begin{figure}[htbp]
  \centering
  \begin{subtable}{0.48\textwidth}
    \centering
    \caption{Ablations of training components.}
    \begin{tabular}{@{}ccc@{}}
    \toprule
    Method& IGD$\downarrow$& GD$\downarrow$\\
    \midrule
     A& 0.0155& 0.0182\\
     B& \textbf{0.0107}&0.0067\\
     C& 0.0149&0.0158 \\
     D& \textbf{0.0107}&0.0069 \\
     Ours& 0.0281&\textbf{0.0006}\\
    \bottomrule
    \end{tabular}
    \label{tab:train_ablation}
  \end{subtable}\hfill
  \begin{subfigure}{0.48\textwidth}
    \centering
    \includegraphics[width=\textwidth]{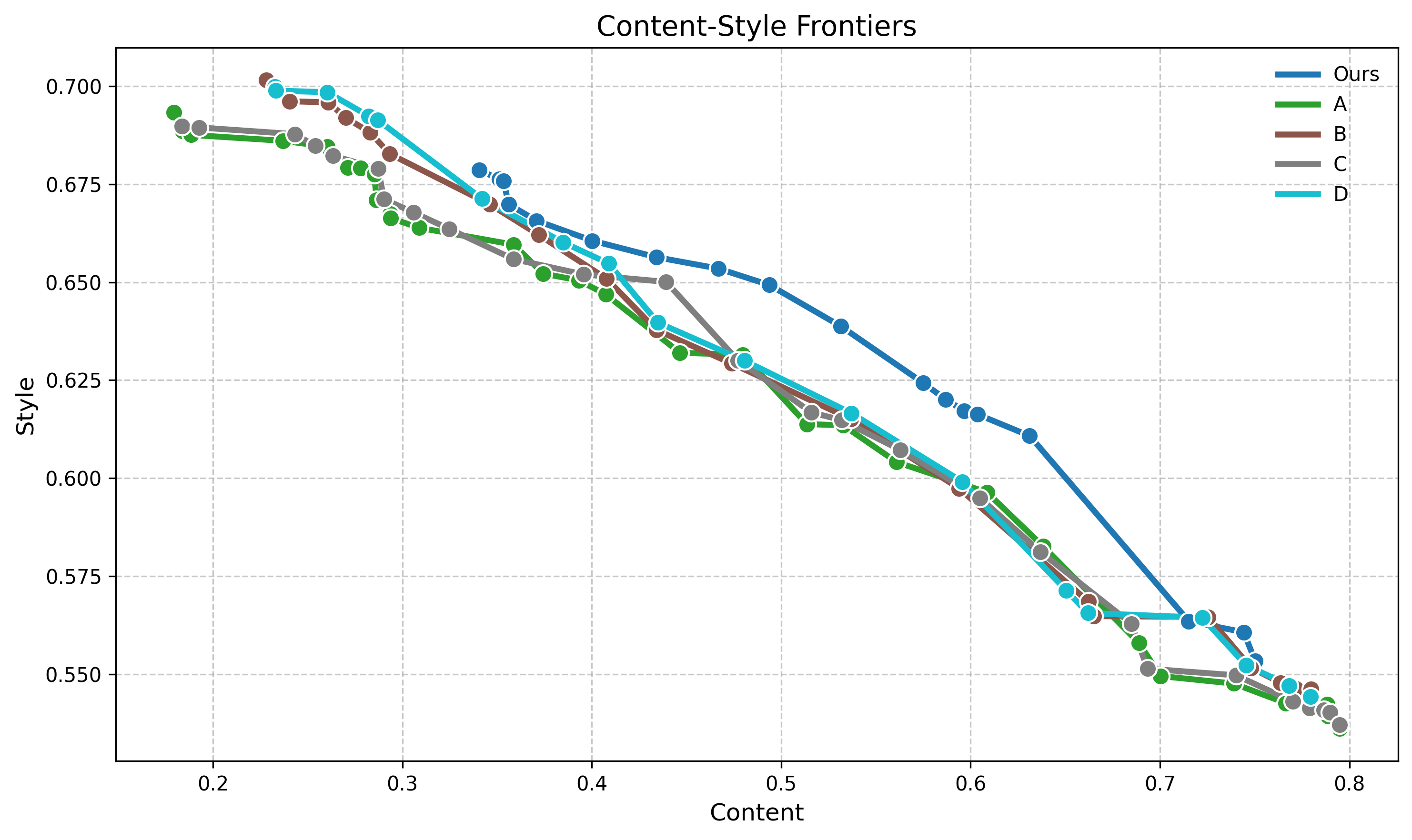}
    \caption{Ablations of training components. Our method outperforms others in terms of approximation accuracy.}
    \label{fig:train_ablation}
  \end{subfigure}
\caption{Ablations of training components.}
\end{figure}

To evaluate the effectiveness of our proposed method, we design several alternative approaches that blend the subspaces of $\mathbf{A}_\text{m}\mathbf{B}_\text{m}$. Specifically, we consider: 
\begin{equation}
\begin{split}
    \Delta \mathbf{W}
&=
\begin{bmatrix}
\mathbf{A}_\text{c} & \mathbf{A}_\text{s}
\end{bmatrix}
\begin{bmatrix}
\mathbf{W}_{11} & \mathbf{O}\\
\mathbf{O} & \mathbf{W}_{22}
\end{bmatrix}
\begin{bmatrix}
\mathbf{B}_\text{c} \\
\mathbf{B}_\text{s}
\end{bmatrix}\text{,}
\end{split}
\label{eq:method_A}
\end{equation}

\begin{equation}
\begin{split}
    \Delta \mathbf{W}
&=
\begin{bmatrix}
\mathbf{A}_\text{c} & \mathbf{A}_\text{s}
\end{bmatrix}
\mathbf{W}
\begin{bmatrix}
\mathbf{B}_\text{c} \\
\mathbf{B}_\text{s}
\end{bmatrix}\text{,}
\end{split}
\label{eq:method_B}
\end{equation}

\begin{equation}
\begin{split}
    \Delta \mathbf{W}
&=
\mathbf{A}_\text{m}
\begin{bmatrix}
\mathbf{I}+(\mathbf{J}-\mathbf{I})\otimes \mathbf{W}_{11} & \mathbf{O}\\
\mathbf{O} & \mathbf{I}+(\mathbf{J}-\mathbf{I})\otimes \mathbf{W}_{22}
\end{bmatrix}
\mathbf{B}_\text{m}\text{,}
\end{split}
\label{eq:method_C}
\end{equation}

\begin{equation}
\begin{split}
    \Delta \mathbf{W}
&=
\begin{bmatrix}
\mathbf{A}_\text{c} & \mathbf{A}_\text{s}
\end{bmatrix}
(\mathbf{I}+(\mathbf{J}-\mathbf{I})\otimes \mathbf{W})
\begin{bmatrix}
\mathbf{B}_\text{c} \\
\mathbf{B}_\text{s}
\end{bmatrix}\text{,}
\end{split}
\label{eq:method_D}
\end{equation}
where $\mathbf{J}$ denotes an all-ones matrix. The learnable parameters $\mathbf{W}_{11},\mathbf{W}_{22}\in \mathbb{R}^{r \times r}$ and $\mathbf{W}\in \mathbb{R}^{2r \times 2r}$ are initialized as identity matrices. We refer to the methods in \cref{eq:method_A}-\cref{eq:method_D} as methods A, B, C, and D, respectively. The quantitative results are summarized in \cref{tab:train_ablation}. Our method achieves superior approximation accuracy compared to all alternatives. Although other variants show a slight edge in the IGD metric, this is primarily due to their coverage of invalid regions with low content similarity (as illustrated in \cref{fig:train_ablation}). In contrast, our method focuses more intently on identifying meaningful optimal solutions, thereby achieving the best approximation accuracy. 

We also conduct ablations on the loss components of the Content-Style Balance loss, as reported in \cref{tab:loss_ablation}. In particular, we define a squared loss component: 
\begin{equation}
    \mathcal{L}_\text{sq}=(\mathcal{L}_\text{c}-\mathcal{L}_\text{s})^2\text{.}
\end{equation}
We compare our Non-linear Content-Style Blending with two alternative blending strategies: Constant Content-Style Blending and Linear Content-Style Blending (\cref{tab:inference_ablation}). These methods are defined as:
\begin{equation}
    \triangle \mathbf{W}
=\begin{bmatrix}
\mathbf{A}_\text{c} & \mathbf{A}_\text{s}
\end{bmatrix}
\begin{bmatrix}
\mathbf{I} & \mathbf{W}_{12}\\
\mathbf{W}_{21} & \mathbf{I}
\end{bmatrix}
\begin{bmatrix}
k\mathbf{B}_\text{c}\\
(1-k)\mathbf{B}_\text{s}
\end{bmatrix}\text{,}
\end{equation}
\begin{equation}
    \triangle \mathbf{W}
=\begin{bmatrix}
\mathbf{A}_\text{c} & \mathbf{A}_\text{s}
\end{bmatrix}
\begin{bmatrix}
\mathbf{I} & \mathbf{W}_{12}\\
\mathbf{W}_{21} & \mathbf{I}
\end{bmatrix}
\begin{bmatrix}
\text{max}((1-kt),0)\mathbf{B}_\text{c}\\
\text{min}(kt,1)\mathbf{B}_\text{s}
\end{bmatrix}\text{.}
\end{equation}

\begin{table}[]
    \centering
    \caption{Ablations of inference components.}
    \begin{tabular}{@{}ccc@{}}
    \toprule
    Method& IGD$\downarrow$& GD$\downarrow$\\
    \midrule
     CCSB& 0.0187& 0.0127\\
     LCSB& 0.0083&0.0044\\
     NCSB& \textbf{0.007}&\textbf{0.0023} \\
    \bottomrule
    \end{tabular}
    \label{tab:inference_ablation}
\end{table}

\begin{table}[]
    \centering
    \caption{Ablations of loss components.}
    \begin{tabular}{@{}cccccc@{}}
    \toprule
    $\mathcal{L}_\text{prior}$&$\mathcal{L}_\text{merger}$&$\mathcal{L}_\text{sq}$&$\mathcal{L}_\text{lbr}$& IGD$\downarrow$& GD$\downarrow$\\
    \midrule
    $\checkmark$&$\checkmark$&$\times$&$\times$ & 0.0211& 0.0074\\
    $\times$&$\checkmark$&$\times$&$\checkmark$ &0.0264 &0.0074\\
    $\checkmark$&$\times$&$\times$&$\checkmark$ & 0.0152&0.0095 \\
    $\checkmark$&$\checkmark$&$\checkmark$&$\times$ &0.0132 & 0.0108\\
    $\checkmark$&$\checkmark$&$\times$&$\checkmark$ & \textbf{0.0112}&\textbf{0.0016}\\
    \bottomrule
  \end{tabular}
    \label{tab:loss_ablation}
\end{table}

\section{Conclusions}
In this work, we propose a new method for achieving a more flexible and precise balance between content and style in image generation using diffusion models. By leveraging Content-Style Subspace Blending and Content-Style Balance loss, we provide continuous control over the content-style trade-off, significantly expanding the content-style frontier. Additionally, we introduce a Non-linear Content-Style Blending strategy, leveraging the property of diffusion models to generate content structure early and style details later. This approach leads to notable improvements in both content preservation and style expression. Experimental results demonstrate that our method outperforms existing techniques in terms of both quantitative metrics (i.e., IGD and GD), and qualitative visual results. 
{
    \small
    \bibliographystyle{ieeenat_fullname}
    \bibliography{main}
}


\end{document}